\documentclass[10pt,twocolumn,letterpaper]{article}

\usepackage[pagenumbers]{cvpr} 

\newcommand{\name}{\textsc{PathEval}\xspace}
\usepackage{soul}

\usepackage{graphicx}
\usepackage{amsmath}
\usepackage{multirow}
\usepackage{multicol}
\usepackage{amssymb}
\usepackage{lineno}
\usepackage{tabularx}
\usepackage{longtable}
\usepackage{float}
\usepackage{amsmath}
\usepackage{subcaption}
\usepackage{tcolorbox}
\usepackage{caption}
\tcbuselibrary{listings,breakable}

\definecolor{cvprblue}{rgb}{0.21,0.49,0.74}
\usepackage[pagebackref,breaklinks,colorlinks,allcolors=cvprblue]{hyperref}

\def\paperID{16841} 
\def\confName{CVPR}
\def\confYear{2025}

\title{Evaluating Vision-Language Models as Evaluators in Path Planning}

\author{
    Mohamed Aghzal$^{\spadesuit}$, Xiang Yue$^{\clubsuit}$, Erion Plaku$^{\diamondsuit}$\thanks{The work by E. Plaku is supported by (while serving at) the National Science Foundation. Any opinion, findings, and conclusions or recommendations expressed in this material are those of the authors and do not necessarily reflect the views of the National Science Foundation.},  Ziyu Yao$^{\spadesuit}$\\
    $^{\spadesuit}$George Mason University, $^{\clubsuit}$Carnegie Mellon University, $^{\diamondsuit}$National Science Foundation \\
    {\tt\small \{maghzal, ziyuyao\}@gmu.edu, xyue2@andrew.cmu.edu, eplaku@nsf.gov}
}
\begin{document}
\maketitle
\begin{abstract}

Despite their promise to perform complex reasoning, large language models (LLMs) have been shown to have limited effectiveness in end-to-end planning. This has inspired an intriguing question: if these models cannot plan well, can they still contribute to the planning framework as a helpful plan evaluator? 
In this work, we generalize this question to consider LLMs augmented with visual understanding, i.e., Vision-Language Models (VLMs).
We introduce \name, a novel benchmark evaluating VLMs as plan evaluators in complex path-planning scenarios. Succeeding in the benchmark requires a VLM to be able to abstract traits of optimal paths from the scenario description, demonstrate precise low-level perception on each path, and integrate this information to decide the better path. 
Our analysis of state-of-the-art VLMs reveals that these models face significant challenges on the benchmark. We observe that the VLMs can precisely abstract given scenarios to identify the desired traits and exhibit mixed performance in integrating the provided information. Yet, their vision component presents a critical bottleneck, with models struggling to perceive low-level details about a path. Our experimental results show that this issue cannot be trivially addressed via end-to-end fine-tuning; rather, task-specific discriminative adaptation of these vision encoders is needed for these VLMs to become effective path evaluators.\footnote{Source code: \href{https://github.com/MohamedAghzal/PathEval}{https://github.com/MohamedAghzal/PathEval}}\footnote{Dataset: \href{https://huggingface.co/datasets/maghzal/PathEval}{https://huggingface.co/datasets/maghzal/PathEval}}

\end{abstract}    
\section{Introduction}
\label{sec:intro}

\begin{figure*}[h!]
\begin{tcolorbox}[colback=cvprblue!5!white, colframe=cvprblue!75!black, title=Example of an instance of \name]
    \label{box:model-response} 
    \begin{minipage}{0.35\textwidth}
        \centering
        \includegraphics[width=\linewidth]{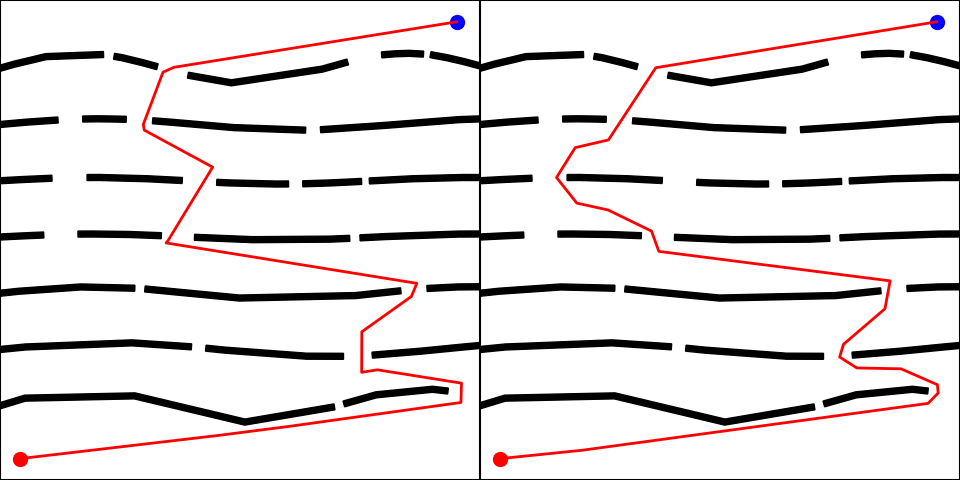} 
        \caption*{2D}
    \end{minipage}
    \hfill
    \begin{minipage}{0.6\textwidth}
        \centering
        \includegraphics[width=\linewidth]{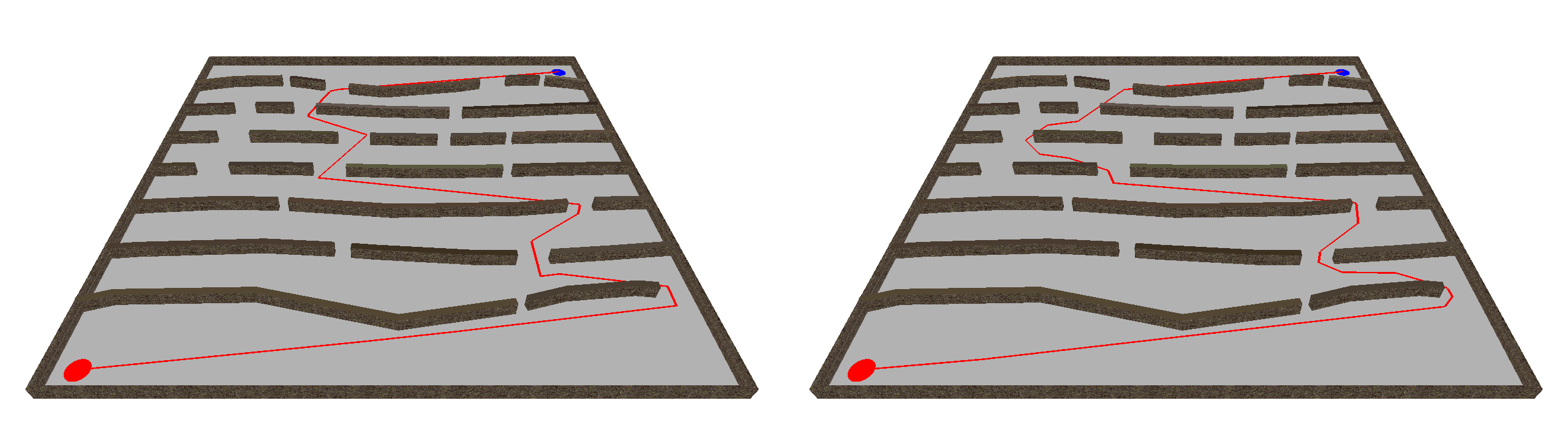}
        \caption*{3D}
    \end{minipage}
    \hfill
    \begin{minipage}{\textwidth}
        \footnotesize
        \rule{\textwidth}{0.4pt}  

        \textbf{Path Descriptors:} \\
\textbf{Path 1:} \textit{Min. clearance:} 1.07, \textit{Max. clearance:} 8.91, \textit{Avg. clearance:} 2.62, \textit{Path length:} 243.46, \textit{Smoothness:} 843.07, \textit{Number of sharp turns:} 4, \textit{Maximum angle:} 124.35. \\
\textbf{Path 2:} \textit{Min. clearance:} 1.57, \textit{Max. clearance:} 8.91, \textit{Avg. clearance:} 3.18, \textit{Path length:} 243.75, \textit{Smoothness:} 809.65, \textit{Number of sharp turns:} 1, \textit{Maximum angle:} 93.12. \\
        \rule{\textwidth}{0.4pt} 
        \textbf{Scenario \#1: } The agent navigating this maze is a large truck, so sharp turns (90 degrees or larger) are harder to make. It should also stay on a straight line (unless it is making a turn) as it is driving through heavy traffic. \\
        \textbf{Required Descriptors: } {\underline{Smoothness (\( \downarrow \))}, \underline{Number of Sharp turns (\( \downarrow \))},  \underline{Maximum angle (\( \downarrow \))}} \\
        \textbf{Ground Truth: } Path 2
        \\
        \textbf{Scenario \#2: } An autonomous firefighting robot is designed to navigate and operate within burning buildings to extinguish fires and rescue trapped individuals. It should explore as much of the area as possible, while maintaining a safe distance from the walls to avoid damage. \\
        \textbf{Required Descriptors: } {\underline{Avg. Clearance (\( \uparrow \))}, \underline{Min. Clearance (\( \uparrow \))},  \underline{Path length (\( \uparrow \))}} \\
        \textbf{Ground Truth: } Path 1\\
        \textbf{Scenario \#3:} As the vehicle is traversing a warzone, it must stay concealed from enemy operatives, making use of covers like walls and avoiding open spaces as much as possible. It should also reach its target as quickly as possible.\\
        \textbf{Required Descriptors: } {\underline{Avg. Clearance (\( \downarrow \))}, \underline{Max. Clearance (\( \downarrow \))},  \underline{Path length (\( \downarrow \))}} \\
        \textbf{Ground Truth: } Path 2\\
        \rule{\textwidth}{0.4pt} 
        \textbf{Prompt} \\ \textit{\textbf{The agent navigating this maze is a large truck, so sharp turns (90 degrees or larger) are harder to make. It should also stay on a straight line (unless it is making a turn) as it is driving through heavy traffic.} Which path better achieves the task? Path 1 is on the left side and Path 2 is on the right side. Your answer should follow the format below:\\
        \textbf{Answer}: Path 1 or Path 2.\\
        \textbf{Explanation}: Why you chose the path (1 or 2).}

        \rule{\textwidth}{0.4pt}  

         \textbf{GPT-4o}\\
        \textit{\textbf{Answer:}} \textcolor{red}{\textbf{Path 1.}}\\
        \textit{\textbf{Explanation: }} Path 1 has \textbf{\textcolor{red}{fewer sharp turns}} and maintains longer straight segments compared to Path 2. This is more suitable for a large truck navigating through heavy traffic, as it minimizes the need for difficult maneuvers and allows for smoother driving. Path 2, on the other hand, \textbf{\textcolor{red}{has more frequent and sharper turns}}, which would be more challenging for a large truck to navigate. \\
        
        \textbf{LLaVA-NeXT-7b} \\
        \textbf{Answer:} \textcolor{red}{\textbf{Path 1.}}\\
        \textit{\textbf{Explanation:}} Path 1 is better suited for a large truck because \textbf{\textcolor{red}{it has fewer sharp turns (90 degrees or larger)}} compared to Path 2. The truck \textbf{\textcolor{red}{would have to make only one 90-degree turn}}, which is easier to execute than \textbf{\textcolor{red}{the two 90-degree turns in Path 2}}. Additionally, Path 1 has more straight sections, which would be more efficient for a large truck to navigate through heavy traffic. The fewer turns and the greater number of straight sections make Path 1 the more suitable route for a large truck. 
    \end{minipage}
\end{tcolorbox}
\caption{Example of \name. The benchmark consists of 14,550 instances of path pairs rendered in both 2D and 3D and mapped to 15 decision-making scenarios. Success on this task is tied to three distinct levels: \textbf{1) Attribute abstraction:} recognizing what aspects make a path ideal, \textbf{2) Low-level perception:} extracting the required attributes for each path from the images, and \textbf{2) Information integration:} synthesizing the collected information to make a decision. We test a set of VLMs on the task and find that they struggle particularly with low-level perception.
Incorrect answers by different models are shown (\textcolor{red}{\textbf{explanations indicating misperception}} are highlighted).} 
\label{fig:example}
\vspace{-0.14in}
\end{figure*}


The impressive capabilities of Large Language Models (LLMs) \cite{brown2020languagemodelsfewshotlearners, caron2021emerging, wei2022chain} and Vision-Language Models (VLMs) \cite{zhang2024multimodalchainofthoughtreasoninglanguage} have led to an increasing interest in applying them to automated motion planning and navigation tasks \cite{pan-etal-2024-langnav, vlp2024}. 
{However, the inherent limitations of these models in 
long-horizon planning have rendered them ineffective as end-to-end motion planners~\cite{chen2023autotamp, aghzal2024look, yang2024guidinglonghorizontaskmotion, aghzal2024can}.}
{This has made researchers wonder: if these models cannot be good motion planners themselves, can they still support a motion planning framework? Intuitively, using these models still holds the promise of significantly enhancing the motion planning framework, as they have learned extensive factual and commonsense knowledge that could benefit planning during their pre-training. As a result, there has been an emerging paradigm exploring how these models can be leveraged in combination with traditional methods \cite{kambhampati2024position}.}

One particularly interesting approach within this line of work involves using these models as \emph{plan evaluators}. Motivated by the intuition that “evaluation is easier than generation”~\cite{leike}, several efforts have explored leveraging these models as “critics” to assess the quality of generated plans \cite{baumli2024visionlanguagemodelssourcerewards, xie2024textreward, zhong2024policyimprovementusinglanguage, guan2024task}. However, most of these efforts have focused on scenarios that require only limited, high-level visual perception, without necessitating fine-grained or precise perceptual abilities. On the other hand, while {there have been prior works similarly investigating VLMs' low-level perception}
\cite{kamath2023whats, tong2024eyeswideshutexploring}, {studies specifically about the use of these models in planning remain limited. Tasks such as motion planning often require fine-grained visual understanding in highly specific contexts, while also drawing on broad commonsense knowledge acquired during pre-training. Hence, there is a pressing need to investigate the potential of VLMs to understand both low-level visual details and leverage these visual signals for high-level reasoning.} 

In this work, we explore whether we can utilize VLMs as evaluators in highly intricate continuous path planning problems. We introduce \name (Figure~\ref{fig:example}), a controllable benchmark designed around path planning in complex environments under diverse commonsense decision-making scenarios (e.g., {A firefighting robot has to cover as much of the area as possible to extinguish fires and scout for survivors, thus needs to prioritize paths with more coverage and higher clearance}). {Traditionally, encoding constraints in a planning algorithm requires significant human effort {in hand-crafting scenario-specific criteria for path evaluation, such that the constraints can be injected into the planner during the search process \cite{Plonski2013, lim2019constrainedpathplanning, petres2007pathplanning}. 
However, this not only is a tedious and costly process but also cannot scale up to handle the countless intricate scenarios in meeting real-life planning needs.} {An effective VLM as a ``reward model'' relieving humans from customizing evaluation criteria for specific scenarios could enable scalable, general-purpose planners that can adapt to diverse commonsense scenarios described in natural language.}}

In \name, a VLM is tasked with comparing two paths within a given decision-making context and selecting the one that better satisfies the constraints outlined by the scenario. Success on this task {requires effective performance across} three distinct levels: 1) \emph{attribute abstraction:} recognizing the attributes that define a favorable path in a particular decision-making context; 2) \emph{low-level perception:} demonstrating precise low-level perception to determine which path performs better based on the given criteria; and 3) \emph{information integration: } integrating and synthesizing the perceived information to produce an answer. 

Using this benchmark, we analyze the performance of 9 state-of-the-art (SOTA) VLMs, including both the closed-source GPT-4o and GPT-4o-mini~\cite{openai2024gpt4o} and 7 different open-source VLMs of different sizes (i.e., LLaVA-NeXT 7b and 13b \cite{liu2024llavanext}, Qwen2-VL-7b \cite{Qwen2VL}, LLaVA-OneVision-7b \cite{li2024onevision}, LLaMA-3.2-11b \cite{dubey2024llama3herdmodels}, and Intern-VL2 8b and 40b \cite{chen2023internvl}). We find that these models struggle with the path evaluation task (e.g., Qwen2-VL-7b achieving only 50.2\% accuracy). However, when providing these VLMs with \emph{verbalized} path specifications, their performance significantly improves (e.g., 74.2\% accuracy for Qwen2-VL-7b), which reveals {a potential vision bottleneck of these VLMs. Our further analysis confirms these models' weakness in low-level perception, especially when they are tasked to perceive the clearance of a path with respect to surrounding obstacles, and this weakness could be more prominent when the environment and the path representation become more complex. We discover the source of this weakness from the vision encoders used by these VLMs, yet simply fine-tuning the VLMs end-to-end with the vision encoders does not address the issue. Rather, our experiments suggest performing task-specific discriminative adaptation of these vision encoders.} 


\section{Related Work}

\subsection{Vision-Language Models}
The outstanding success of decoder-only LLMs \cite{brown2020languagemodelsfewshotlearners, openai2024gpt4technicalreport, touvron2023llama2openfoundation} has driven the development of Vision-Language Models (VLMs), which extend LLMs with a vision component in an attempt to generalize their performance into a multi-modal setting \cite{liu2023llava, liu2024improvedbaselinesvisualinstruction, Qwen2VL, geminiteam2024geminifamilyhighlycapable, chen2023internvl, li2024onevision, dubey2024llama3herdmodels}. 
VLMs are designed for tasks that require unifying visual and linguistic representations (e.g. visual question answering \cite{vqa2015}). Typically, this is achieved by connecting a vision encoder to a language decoder and projecting the representations into a shared space, allowing visual elements and linguistic components to be linked in a semantically meaningful manner. One of the earliest successful models at achieving this was the CLIP encoder \cite{radford2021learning}, which was trained using contrastive learning to learn representations that map images and their corresponding textual descriptions. Several varieties of CLIP have then been introduced \cite{zhai2023sigmoid, schuhmann2022laionb, OpenCLIP, yu2022coca}. These models, while showing tremendous promises have shown several limitations when it comes to visual tasks \cite{shao2023investigatinglimitationclipmodels}; as such several works \cite{wysoczanska2023clipdino, tong2024eyeswideshutexploring} have sought ways to improve such representations by combining them with vision foundation models such as DINO \cite{caron2021emerging, oquab2023dinov2}. {In this work, we contribute to the research of VLMs with a new benchmark, namely \name, focusing on evaluating VLMs as evaluators for path planning. This benchmark tests VLMs seamlessly on their commonsense understanding (i.e., being able to abstract critical concepts expressed in the described planning scenarios), low-level perception (i.e., precisely perceiving details about paths in complex environments), and the ability to reason about the collected textual and visual information for decision-making.}


\subsection{Automated Planning with VLMs}
Vision-language planning promises more flexible planning frameworks and enhanced human-AI interaction. Therefore, designing systems that can effectively understand natural language instruction and leverage perceptual inputs to conduct planning tasks has been a topic of interest in recent years \cite{tan-etal-2019-learning-nav, reinforced2019, Gu_vln_survey_2022}. The rise of VLMs has led many to investigate the use of these models as vision-language planning agents \cite{driess2023palme, pan-etal-2024-langnav, du2024video, vlp2024}. However, existing literature highlights the limitations of LLMs in spatial reasoning \cite{aghzal2024can, yamada2024evaluating, kamath2023whats} as well as long-horizon planning \cite{valmeekam2024llmscantplanlrms, aghzal2024look, valmeekam2022llmscantplan}. This pushed researchers to explore alternative ways to incorporate VLMs and LLMs into planning frameworks more reliably \cite{silver2023generalizedplanningpddldomains, kambhampati2024position}. One potential direction is their use as plan evaluators, either through the generation of reward functions \cite{xie2024textreward, li2023auto, han2024generatingevolvingrewardfunctions}, or by using them directly as off-the-shelf critics \cite{guan2024task, baumli2024visionlanguagemodelssourcerewards, zhong2024policyimprovementusinglanguage}. The success of such frameworks assumes perfect perception and that the models can accurately perceive visual information and reason about it in order to produce an answer; nevertheless, it has been shown that the representations used by these models fail in highly intricate visual settings \cite{tong2024eyeswideshutexploring}. {Several works have explored the use of VLMs as well as LLMs for path and motion planning \cite{ding2023task, aghzal2024look, aghzal2024can, meng2024llmalargelanguagemodel, chen-etal-2024-mapgpt, yang2024guidinglonghorizontaskmotion, chen2023autotamp, cao_2024_CVPR}, however, to the knowledge of the authors there is no work that explores the use of VLMs as path critics in this context.} Accordingly, we aim to evaluate the ability of VLMs to serve as evaluators in cases requiring navigation in complex environments while adhering to decision-making constraints specified in natural language.

\subsection{Vision-Language Model Benchmarks}

The introduction of multimodal models has prompted the development of several benchmarks that are capable of assessing the performance of these models on visual reasoning tasks such as visual question-answering datasets \cite{zhu2016cvpr, vqa2015, Goyal_2017_CVPR, gqa, Marino_2019_OKVQA, gao2022cric, lu2024mathvista}. However, the rise of foundation models has produced the need for a more holistic evaluation of the perceptual and reasoning capabilities of large VLMs, leading to benchmarks such as MM-Vet \cite{yu2024mm}, MM-Bench \cite{eccv24MMBench} , MMMU \cite{yue2023mmmu, yue2024mmmu, onohara2024jmmmujapanesemassivemultidiscipline, zhang2024cmmmuchinesemassivemultidiscipline} and OmniBench \cite{li2024omnibenchfutureuniversalomnilanguage}. Several benchmarks have also specifically been designed to assess the perception capabilities of these models and explore the limitations associated with visual hallucinations and optical illusions \cite{guan2023hallusionbench, Li-hallucination-2023, tong2024eyeswideshutexploring, sun2024explorehallucinationlowlevelperception, chen2024multiobject}. Our proposed benchmark provides a flexible yet challenging framework for interleavedly assessing the low-level perception and reasoning capabilities of VLMs.

\section{The \name Benchmark}
\label{sec:dataset}

Motivated by the need to evaluate VLMs as path evaluators in real-world planning scenarios, we introduce \name, a controllable and extensible benchmark focused on \emph{path planning}~\cite{lozano1979pathplanning, LaValle1998RapidlyexploringRT, kuffner-RRTConnect, kavraki1996prm} in complex environments under a diverse set of decision-making constraints. We list the decision-making scenarios as well as the descriptors they attempt to optimize in Appendix \ref{sec:appendix-scenarios}. {In total, \name includes 14,550 tasks over more than 1,150 distinct environments and 15 distinct scenarios.} Below, we introduce a formal description of the task and the dataset construction. 

\subsection{Task Formulation}

Given two paths, \( P_1 \) and \( P_2 \), and a scenario \( S \), the objective is to determine which path better satisfies the scenario's optimization criteria. Each scenario \( S \) is a high-level description that aims to optimize over a set of \emph{path descriptors} (or \emph{metrics}) \( \mathcal{M} = \{m_1, m_2, \dots, m_k\} \),
where each descriptor \( m_j: P \to \mathbb{R} \) evaluates a specific property of a path (e.g., length, smoothness, or proximity to obstacles). A VLM {\( \mathcal{V} \)} is presented with two images presenting \( P_1 \) and \( P_2 \) in the same environment, respectively.
The model must then decide which path better satisfies the scenario's criteria. To explore the sensitivity of VLMs to the way how a path is presented, \name includes both the 2D and 3D images of the path illustration. {The model is also prompted to generate an explanation to justify its choice.}

\subsection{Environment and Path Generation}

\textbf{Environment Generation:} An environment, as shown in Figure~\ref{fig:example}, is defined by a set of walls \( \mathcal{O} = \{O_1, O_2, \dots, O_n\} \), where each wall \( O_i \) represents an obstacle in the 2D space. 
Each wall is a closed geometric shape described by its vertices, and the set \( \mathcal{O} \) forms the obstacles that the path must avoid. In this work, we leverage the environments of Plaku et. al. \cite{Plaku_Plaku_Simari_2018}, which{ consists of {four} types of obstacle arrangements: (1) \emph{rings,} where the environments are structured as mazes with circular walls, (2) \emph{waves,} which consist of wavy horizontal obstacles, (3) \emph{mazes,} which consist of both vertical and horizontal walls forming a complex maze structure, as well as {(4)} \emph{random}, which consist of randomly placed obstacles.}

\vspace{1mm}
\noindent\textbf{Path Synthesis via the Randomly-exploring Rapid Tree (RRT) algorithm:} To generate path candidates in \name, we leverage the RRT path planning algorithm~\cite{LaValle1998RapidlyexploringRT}. 
{Starting from the initial location in the environment, the algorithm works by building a tree which expands itself by randomly selecting the next location in the environment while avoiding obstacles, until it reaches the goal.}
In this work, we use the Open Motion Planning Library (OMPL) \cite{ompl2012} and implement the RRT-Connect algorithm~\cite{kuffner-RRTConnect}. We note that while we use RRT in the current benchmark, our codebase is adaptable and can incorporate most path-planning algorithms provided by OMPL. We encourage future research building on our benchmark to experiment with other algorithms as well. 

\subsection{Path Descriptors}\label{subsec:path-descriptors}
We collect the following descriptors $\mathcal{M}$ for each of a generated path: \textbf{\emph{Minimum Clearance}} measures the smallest distance between any point on the path and the nearest obstacle; \textbf{\emph{Maximum Clearance}} measures the largest distance between any point on the path and the nearest obstacle; \textbf{\emph{Average Clearance}} computes the average distance between all points on the path and the nearest obstacle; \textbf{\emph{Path Length}} is calculated by summing up the Euclidean distances between consecutive points on the path; \textbf{{\emph{Smoothness}}} is defined as the sum of the angles between consecutive segments of the path, measuring how smoothly the path changes direction; \textbf{\emph{Number of Sharp Turns}} counts the number of turns in the path where the angle between consecutive segments exceeds 90 degrees; and \textbf{\emph{Maximum Angle}} denotes the largest angle between any two consecutive segments of the path. 
{The three Clearance metrics and Path Length share the same measuring unit, i.e., one grid size; Smoothness and Maximum Angle are measured by degree; and the Number of Sharp Turns is an integer count.}
{We include the formula of each descriptor in Appendix~\ref{app:descriptors}.}

\subsection{Natural Language Descriptions of Scenarios}

To create a sufficiently challenging path-planning evaluation benchmark, we design a total of 15 decision-making scenarios that aim to optimize different combinations of the path descriptors.
{For instance, Scenario \#2 (as shown in Figure \ref{fig:example}) requires searching through an area affected with fire in search for survivors, and the agent thus must cover as much ground as possible. In contrast, Scenario \#3 indicates that the path is to be executed within a warzone; as a result, the vehicle has to remain hidden and take the shortest route. As such, given the same set of paths, the one minimizing the path length is favored by Scenario \#3 while Scenario \#2 needs to maximize this value.}
A complete summary of the 15 scenarios, along with the descriptors each scenario aims to optimize, is presented in Tables~\ref{tab:scenarios}-\ref{tab:scene-metrics} in Appendix~\ref{sec:appendix-scenarios}.

\subsection{Task Pairs Selection}


For each environment we synthesized, we ran the RRT planner 30 times to generate different paths. Upon eliminating paths that did not reach the goal, we selected path pairs that exhibited the greatest dissimilarity in terms of path descriptors. 
{Specifically, we first represented each path using a (7-dimension) vector of its path descriptor values. Given that each path descriptor ranged in a dramatically different scale, we normalized the vector by performing the min-max scaling, i.e., scaling each value \( x \) in the vector to \( x' = \frac{x - x_{\text{min}}}{x_{\text{max}} - x_{\text{min}}} \), where \( x \) represents each value in the vector, and \( x_{\text{min}} \) and \( x_{\text{max}} \) are the minimum and maximum values of each descriptor across paths sampled from the same environment, respectively. We then measured the distance between two paths by calculating the Euclidean distance between their normalized path descriptor vectors, and selected 5 path pairs with the largest distances. Each path is included in only one pair to avoid redundancy. We repeated the same selection procedure for a total of 1,150 environments.}

Upon performing this generation, we matched each pair with all fifteen scenarios; however, we only kept pairs where there was a significant difference in at least one of the descriptors required for the particular scenario. {In other words, we ensure that the absolute difference is greater than a predefined threshold (0.8 for clearance descriptors, 50 for path length, 90 for smoothness, 1 for the number of sharp turns, and 30 for maximum angle) for at least one of the required descriptors}. This 
makes it more likely that the difference is more noticeable to the naked eye, and thus the paths can be compared by visual inspection.

We constructed our final evaluation benchmark by randomly selecting 70 task pairs from each scenario, resulting in 1,050 pairs in total. The remaining task pairs (a total of 13,500) are used as the training set to facilitate the fine-tuning experiments in Section~\ref{sec:finetuning}.
\section{Can SOTA VLMs Evaluate Planned Paths?}
\label{sec:preliminary}

In order for VLMs to perform successfully in our benchmark, they need to perform successfully at three different levels, i.e, recognizing the critical descriptors required by each scenario (\textbf{Attribute Abstraction}), exhibiting sharp low-level perception to precisely assess each path's properties (\textbf{Low-level Perception}), and integrating the prior information to make a rational decision on the better path (\textbf{Information Integration}). {Among them, the first two levels reflect parallel properties that serve as a foundation for the third level.}
In this section, we evaluated a set of 9 VLMs on \name in a zero-shot manner and analyzed their capabilities at these three levels. These VLMs include (1) two closed-source VLMs, i.e., GPT-4o and GPT-4o-mini~\cite{openai2024gpt4o},
and (2) seven open-source VLMs with various sizes, including LLaVA-NeXT-7b and LLaVA-NeXT-13b \cite{liu2024improvedbaselinesvisualinstruction, liu2024llavanext}, LLaVA-OneVision-7b \cite{li2024onevision}, Qwen2-VL-7b \cite{Qwen2VL}, LLaMA-3.2-11b \cite{dubey2024llama3herdmodels}, and Intern-VL2-8b and Intern-VL2-40b \cite{chen2023internvl}. {We include all prompt scripts used in this Section in Appendix~\ref{app:prompts}.}

\subsection{Overview of VLMs on \textbf{\name}}\label{sec:overall}

\begin{table*}[tb]
    \centering
    \footnotesize
    \begin{tabular}{lccccc}
        \toprule
        \multirow{2}{*}{\textbf{Model}} & \multicolumn{2}{c}{\textbf{2D}} & \multicolumn{2}{c}{\textbf{3D}} & \multirow{2}{*}{\textbf{Desc Only}} \\
        \cmidrule(r){2-3} \cmidrule(r){4-5}
        & \textbf{\name} & \textbf{\name w Desc.} & \textbf{\name} & \textbf{\name w Desc.} & \\
        \midrule
        \textbf{GPT-4o-mini} & 0.520 & 0.750 & 0.508 & 0.745 & 0.680 \\
        \textbf{GPT-4o} &\textbf{0.665} & \textbf{0.860} & \textbf{0.624} & \textbf{0.895} & \textbf{0.894} \\
        \midrule
        \textbf{LLaVa-NeXT-7b} & 0.501 & 0.524 & 0.499 & 0.517 & 0.514 \\
        \textbf{Qwen2-VL-7b} & 0.502 & \textbf{0.731} & \textbf{0.511} & \textbf{0.742} & \textbf{0.737} \\
        \textbf{LLaVA-OneVision-7b} & 0.505 & 0.718 & 0.509 & 0.739 & 0.721 \\
        \textbf{Intern-VL2-8b} & 0.489 & 0.654 & 0.505 & 0.691 & 0.648 \\
        \textbf{LLaMa-3.2-11b} & 0.480 & 0.695 & 0.460 & 0.680 & 0.686 \\
        \textbf{LLaVa-NeXT-13b} & \textbf{0.509} & 0.620 & 0.494 & 0.601 & 0.630 \\
        \textbf{Intern-VL2-40b} & 0.506 & 0.688 & 0.496 & 0.717 & 0.679 \\
        \midrule
        \textbf{Random Baseline} & 0.500 & 0.500 & 0.500 & 0.500 & 0.500 \\
        \bottomrule
    \end{tabular}
    \caption{Accuracy of VLMs on \name based on 2D and 3D environment images. To investigate the potential vision bottleneck of VLMs, we additionally present each model's accuracy when explicit descriptor values are provided in the language prompts (``\name w Desc.''). The last column (``Desc Only'') shows the model performance when only the textual descriptor values are provided.}
    \label{tab:overall_accuracy}
    \vspace{-0.1in}
\end{table*}

The performance of the 9 VLMs on \name is showcased in Table~\ref{tab:overall_accuracy}. We notice that all of the models, except GPT-4o, fail to perform significantly better than a simple random baseline, indicating significant limitations. 
For GPT-4o, we also notice a 4\% higher accuracy on 2D images compared to prompting with 3D inputs.
This observation indicates that the model is prone to visual illusions introduced by the 3D images{, when it has to rely on solely the image for decision-making (although we observe an opposite effect of 2D vs. 3D when verbalized descriptor values are provided)}. In the remaining section, we will further break down these models' capabilities to gain a deeper understanding of their failure on this task.


\vspace{1mm}
\noindent\textbf{Providing verbalized path information yields better task accuracy, implying reasonable VLM performance in attribute abstraction.}
In Table~\ref{tab:overall_accuracy}, we further show the performance of each model when we explicitly list the value for each descriptor as part of the language prompt (i.e., ``\name w Desc.''). We notice a {11.1\%-27.1\%} improvement across most models, indicating that when given low-level details, the models can better filter out the information and make better comparisons. This points out that the bottleneck for these VLMs's better performance lies in their inability to accurately perceive low-level information about the paths (we discuss this in more detail in Section~\ref{sec:level2}), whereas these models generally have a reasonable capability in abstracting the critical attributes for decision making in various scenarios. {In Appendix~\ref{app:level1-analysis}, we include an experiment where we explicitly query each VLM to identify the critical path metrics for each scenario; the result corroborates our hypothesis.} {In particular, we find that for most models, the success for identifying a required descriptor is over 92\%.} {Finally, a surprising observation happens to LLaVa-NeXT-7b. We notice that this model suffers particularly severely from hallucination; even when the textual descriptor values are provided and when the model can correctly identify critical path metrics based on our analysis in Appendix~\ref{app:level1-analysis}, it fails to pick the better paths. We show an example of its explanation when textual descriptors are provided in Figure~\ref{fig:example-llava-desc} of Appendix~\ref{app:error-analysis}.}

\noindent \textbf{VLMs exhibit mixed performance in integrating visual and textual path information.}
We take a further look into the model performance when both the image and the textual descriptor values are provided, and contrast it with their performance when only the textual path descriptions are provided (i.e., ``Desc Only''). Interestingly, we observe mixed information integration performance from these models. For GPT-4o, Qwen2-VL-7b, LLaVA-OneVision-7b, LLaMa-3.2-11b, and LLaVa-NeXT-13b, their performance based on only descriptor values has no obvious difference to their best performance when (2D or 3D) images are also provided. This observation implies that these models do not benefit from the additional image information when the textual path descriptions are provided. Instead, sometimes the images (e.g., 2D images for GPT-4o and 3D images for LLaVa-NeXT-13b) may confuse their understanding of the textual descriptors, resulting in a worse accuracy compared to Desc Only. For GPT-4o-mini, Intern-VL2-8b, and Intern-VL2-40b, however, providing both the visual and textual path information offers $\sim$4\%-7\% performance gain over Desc Only, indicating better information integration from these VLMs.

In Appendix~\ref{app:gpt4o-breakdown}, we present a breakdown of GPT-4o's performance by scenarios. We show that performance varies greatly from one scenario to the other. Interestingly, we observe that GPT-4o may \emph{overuse} their commonsense knowledge. For instance, in the case of Scenario \#2 shown in Figure~\ref{fig:example}, where the agent is required to maximize the {path length for better }coverage, GPT-4o still favors the shorter path. {This scenario represents a \emph{counterfactual} situation as models are often trained or instructed to seek the shortest paths. Evaluating VLMs in such counterfactual scenarios allows us to effectively probe their task understanding and reasoning, which we consider to be an important direction for future research.}

\begin{table}[tb]
    \footnotesize
    \centering
    \begin{tabular}{lccc}
    \toprule
    \textbf{Model} & \textbf{Default} & \textbf{Flipped} &  \textbf{Random IDs} \\ \toprule
    \textbf{GPT-4o-mini}  & 34/1016 & 22/1026 & 94/956 \\ 
      \textbf{GPT-4o}  & 278/772 & 258/792 & 291/759 \\ 
      \midrule
      \textbf{LLaVa-NeXT-7b}  & 1028/22 & 1041/9 & 580/470 \\ 
       \textbf{Qwen2-VL-7b} & 21/1029 & 130/920 & 127/923 \\
        \textbf{LLaVA-OneVision-7b} & 438/612 & 440/610 & 262/788  \\
        \textbf{Intern-VL2-8b} & 433/610 & 430/620 & 418/632 \\
        \textbf{LLaMA-3.2-11b*} & 601/251 & 620/264 & 731/319 \\
      \textbf{LLaVa-NeXT-13b} & 795/255 & 939/111 & 305/745 \\ 
      \textbf{Intern-VL2-40b} & 394/656 & 410/639 & 510/540\\ 
      \midrule
        \textbf{Ground truth labels} &  & 530/520 & \\
      \bottomrule
    \end{tabular}
    \caption{Performance on the 2D case (\#of times {first path} is chosen / \#of times {second path} is chosen) when we flip the path order or replace their default names with random IDs. *\footnotesize{There are several cases where LLaMA-3.2-11b does not follow the required format and/or does not give an answer, we omit those cases from this table.}}
    \label{tab:faithfulness}
    \vspace{-3em}
\end{table}

\noindent\textbf{The reasoning of VLMs can be unreliable. } 
Limited by the vision bottleneck, we noticed these VLMs fabricating seemingly plausible explanations to justify their path evaluation, despite the fact that they could not actually perceive the necessary path details to perform the task. This fabrication echos findings from other recent work, where LLMs were shown to produce reasoning traces that do not accurately reflect the process of how the models reach an answer~\cite{agarwal2024faithfulnessvsplausibilityunreliability, tupin2024llmsdontsay}. To gain further insights, we performed an analysis comparing 
model performance on \name with cases that consist of: 1) \textbf{flipping the order of paths in the pair}, and 2) \textbf{assigning random IDs to the paths} (e.g. instead of referring to them as ``Path 1'' and ``Path 2'', we use a random sequence such as ``Path Xu2q'' and ``Path fP48''). The results presented in Table~\ref{tab:faithfulness} showcase that VLMs demonstrate bias for a particular label, when they actually do not have the capability to resolve the task. For example, when no matter the default or the flipping labels are used, {LLaVA-NeXT-7b} consistently selects \emph{Path 1} 98\% of the time and fabricates incorrect observations of the two paths in its explanations (Figure~\ref{fig:example}) to support this choice. {As we discussed earlier, LLaVA-NeXT-7b is particularly prone to hallucination in explanations, leading to its random-guess performance with or without textual descriptors.} 
Introducing random IDs as path names mitigates this bias for LLaVa-NeXT-7b (although the model still obtains a close-to-random accuracy on \name) but does not seem to help other models dramatically.

\subsection{What Can Vision-Language Models See?}\label{sec:level2}

The previous subsection clearly highlights the vision component as the bottleneck for path evaluation on \name. In this subsection, we conduct an analysis of the visual perception capabilities of VLMs. We focus our analysis on GPT-4o since it is the only model that performs substantially better than the random baseline in the case where no descriptors are required. 
\begin{table}[t!]

    \centering
    \footnotesize
    \resizebox{0.95\linewidth}{!}{
    \begin{tabular}{lcccc}
    \toprule
    \textbf{Descriptor} & \textbf{Test Set} & \textbf{\( \epsilon_1 \)} & \textbf{\(\epsilon_2\)} & \textbf{\(\epsilon_3\)} \\ \midrule
    Min. Clearance & 0.46/0.46 & 0.50/0.46 & 0.74/0.70 & 0.86/0.74 \\ 
    Max. Clearance & 0.44/0.46 & 0.41/0.49 & 0.46/0.55 & 0.50/0.60 \\ 
    Avg. Clearance & 0.53/0.55 & 0.50/0.52 & 0.70/0.57 & 0.73/0.60 \\ 
    Path Length & 0.58/0.70 & 0.86/0.91 & 0.92/0.86 & 0.94/0.94 \\ 
    Smoothness & 0.74/0.72 & 0.86/0.82 & 0.90/0.90 & 0.90/0.89 \\ 
    \# of Sharp Turns & 0.76/0.73 & 0.80/0.84 & 0.77/0.70 & 0.60/0.55 \\ 
    Max. Angle & 0.71/0.70 & 0.82/0.84 & 0.86/0.88 & 0.94/0.96 \\ 
    \bottomrule
    \end{tabular}
    }
    \caption{GPT-4o's fine-grained perception accuracy (2D/3D) on the test set of \name and three additionally synthesized datasets with increasing metric differences.}
    \label{tab:finegrained-perception}
    \vspace{-0.1in}
\end{table}

\vspace{1mm}
\noindent\textbf{GPT-4o can perceive paths that are significantly different.} In order to get a better understanding of the model's perception capability, we break down its performance in terms of perception by individual metric. Specifically, we prompt GPT-4o to select which path in the pair provides a lower value on each individual metric and report its accuracy. We perform this analysis on both the task pairs in the test set of \name and an additionally synthesized dataset consisting of task pairs with three levels (denoted as $\epsilon_1$, $\epsilon_2$, and $\epsilon_3$) of increasing differences in their descriptor values. We describe the data generation process for this dataset in more detail in Appendix~\ref{app:gen-finegrained}. 



The results for both experiments are shown in Table~\ref{tab:finegrained-perception}. Upon evaluating the performance on the test set of \name by individual metrics, we notice that GPT-4o particularly struggles with Clearance metrics. These metrics typically require a lower level of perception and is naturally more challenging to discriminate than other metrics. 
On the other hand, Smoothness, Number of Sharpest Turns, and Max Angle appear to be easier for the model to capture. {We also note that GPT-4o perceives the path length much more easily in a 3D environment presentation.}
Furthermore, GPT-4o's performance increases as we increase the descriptor difference between paths. 
This, however, is not the case for the Number of Sharp Turns. 
We conjecture that when we increase the number of sharp turns, we also enforce the challenge of ``counting'' the number of satisfying turns, which VLMs have been shown with limitations~\cite{2023GPT4VisionSC}.

\begin{figure}[t!]
    \centering
    \begin{subfigure}{0.25\linewidth}
        \includegraphics[width=\linewidth]{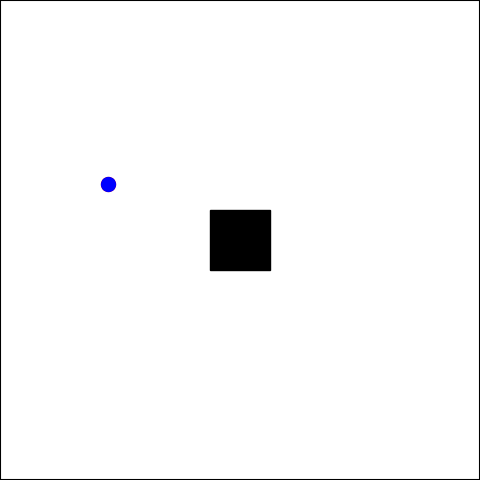}
        \label{fig:point-only}
    \end{subfigure}
    \hfill
    \begin{subfigure}{0.25\linewidth}
        \includegraphics[width=\linewidth]{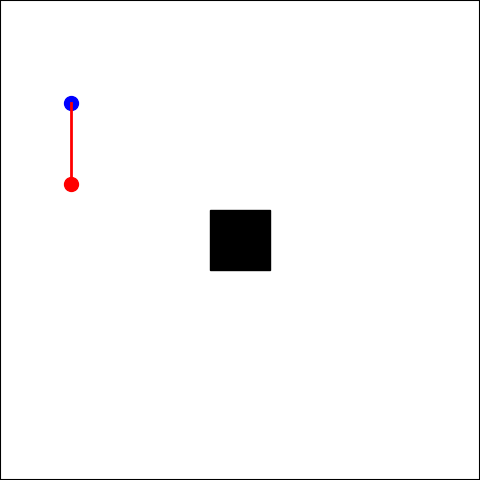}
        \label{fig:straight-line}
    \end{subfigure}
    \hfill
    \begin{subfigure}{0.25\linewidth}
        \includegraphics[width=\linewidth]{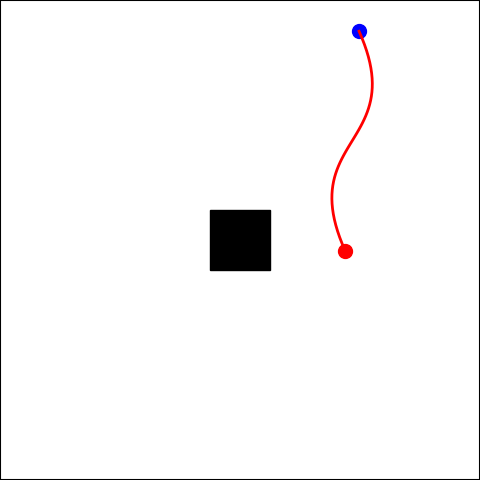}
        \label{fig:curves}
    \end{subfigure}

    \begin{subfigure}{0.25\linewidth}
        \includegraphics[width=\linewidth]{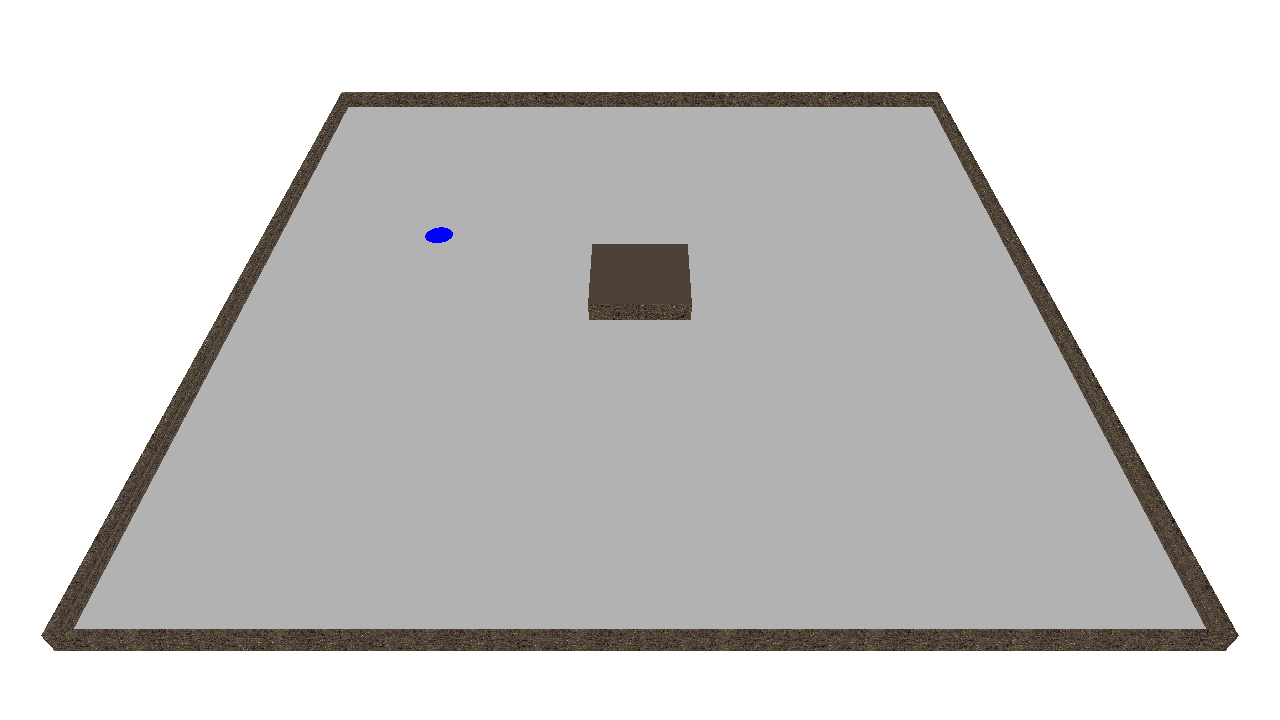}
        \caption{Point}
        \label{fig:point-only}
    \end{subfigure}
    \hfill
    \begin{subfigure}{0.25\linewidth}
        \includegraphics[width=\linewidth]{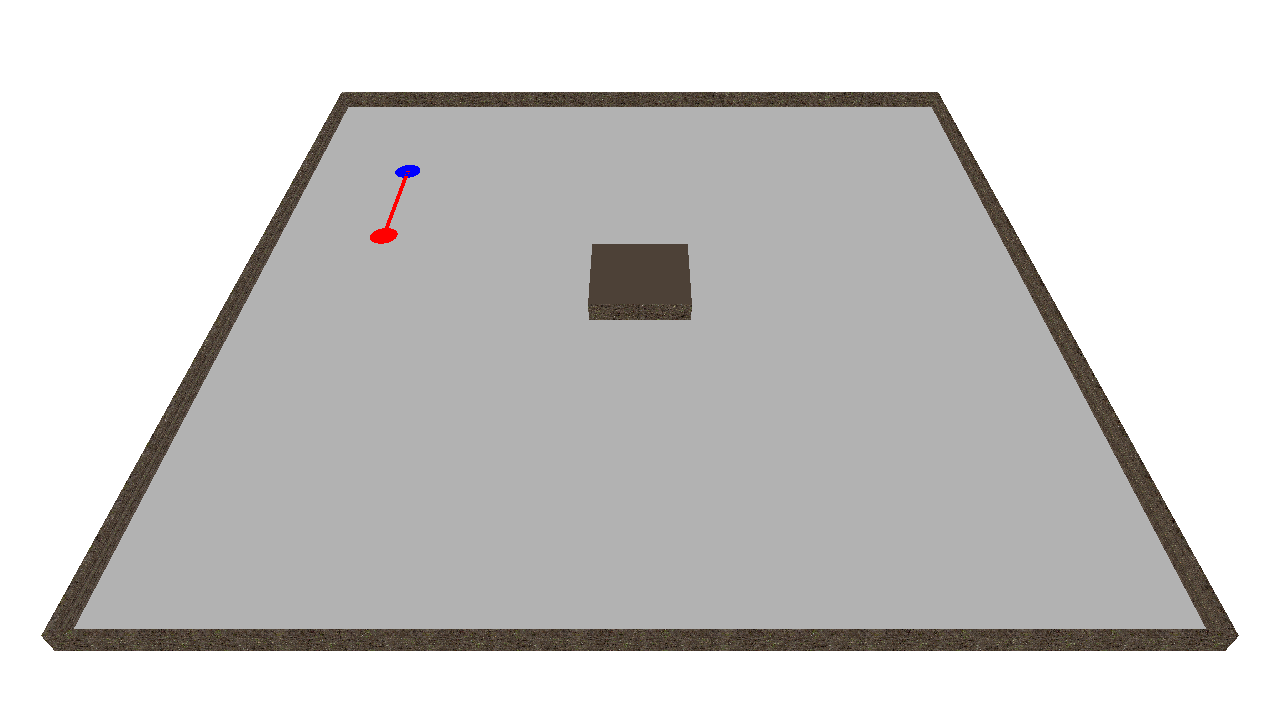}
        \caption{Straight line}
        \label{fig:straight-line}
    \end{subfigure}
    \hfill
    \begin{subfigure}{0.25\linewidth}
        \includegraphics[width=\linewidth]{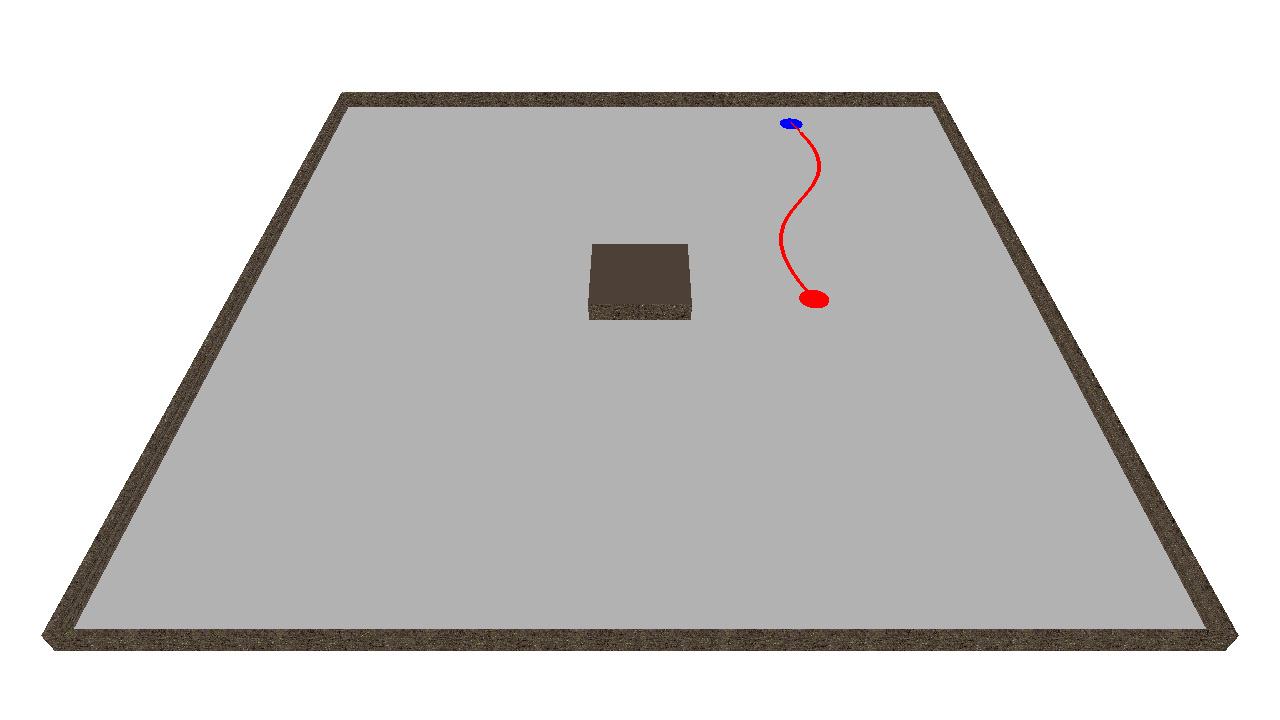}
        \caption{Curve}
        \label{fig:curves}
    \end{subfigure}

\begin{subfigure}{\linewidth}
    \centering
    \includegraphics[width=0.9\linewidth, height=2cm]{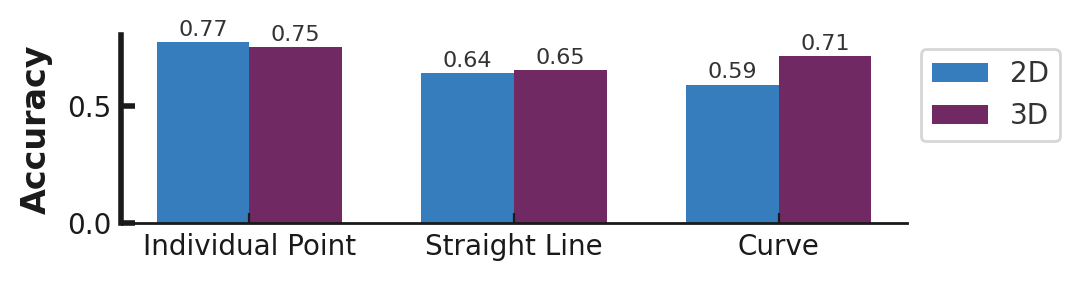}
    \caption{GPT-4o performance on the distance to obstacle under different segment complexities.} 
    \label{fig:segments}
\end{subfigure}

\caption{Example segment complexity test cases in simplified environments and performance across the various settings.}
\label{fig:combined}
\vspace{-3em}
\end{figure}

\vspace{1mm}
\noindent\textbf{Does segment complexity affect performance?} 
We look into whether the complexity of the path segment is the key reason for GPT-4o's limited perception of clearance.
To this end, we test GPT-4o on segments of varying complexity (i.e., points, straight lines, and curves), in a simplified environment with only one rectangular obstacle at the center (Figure~\ref{fig:combined}), and evaluate its accuracy in identifying segments that are closer to the obstacle. For individual points, the clearance is defined by the perpendicular distance from the point to the obstacle; for straight lines and curves, we consider a path closer if one of its endpoints is closer to the obstacle.

{For each segment type, we synthesize 100 pairs by first randomly generating 200 segments and then pairing each with the segment with the greatest distance difference from the obstacle relative to it (i.e., maximizing the absolute difference between the distances of the two segments from the obstacle). This increases the likelihood that the distance is significant enough to be perceivable.} {The average {difference in the clearances of the pairs of segments}
are 14.76, 14.42, and 14.28 for points, lines, and curves respectively.}

The results in Figure~\ref{fig:segments} show that GPT-4o can perform better in very easy scenarios; however, it struggles more as the segment complexity increases. {For instance, the model was able to identify the closer points in 77\% of the cases; however, when considering straight and curved lines, its performance drops to 64\% and 59\%, respectively. A surprising observation is that, in the case of curved lines, GPT-4o's performance is dramatically better in 3D images}. 
{As shown in Appendix~\ref{app:gen-finegrained}, the average Clearance difference of path pairs in \name is merely 0.12 -- 1.31. As the paths in \name are much more complicated than the curves in this experiment, it is expected that GPT~-4o exhibits difficulty in judging paths' clearances. The complexity of the environments (compared to a single square obstacle) could add challenges.}

\section{Fine-tuning a Path Evaluator}\label{sec:finetuning}

\begin{table*}[tb!]
    \footnotesize
    \centering
    \resizebox{0.72\linewidth}{!}{
    \begin{tabular}{lcccc}
        \toprule
        \multirow{2}{*}{\textbf{Model}} & \multicolumn{2}{c}{\textbf{Frozen}} & \multicolumn{2}{c}{\textbf{Fine-tuned}} \\
        \cmidrule(r){2-3} \cmidrule{4-5}
        & \textbf{Accuracy} & \textbf{Avg. Cosine Similarity} & \textbf{Accuracy} & \textbf{Avg. Cosine Similarity} \\
        \midrule
        clip-vit-base-patch32 & 0.510 & 0.914 & 0.783 & 0.514 \\
        clip-vit-large-patch14-336 & 0.498 & 0.907 & 0.749 & 0.548 \\
        OpenCLIP-vit-B-32 & 0.540 & 0.883 & 0.743 & 0.475 \\
        siglip-base-patch16-224 & 0.529 & 0.895 & 0.731 & 0.612 \\
        dino-vit-16 & 0.495 & 0.911 & 0.763 & 0.754 \\
        dino-v2-base & 0.510 & 0.761 & 0.721 & 0.681 \\
        \bottomrule
    \end{tabular}
    }
    \caption{Probing accuracy and average cosine similarity between distinct path images (2D) when the vision encoder is frozen or fine-tuned.
    }
    \label{tab:encoder_comparison}
\end{table*}

\begin{table}[tb!]
    \footnotesize
    \centering
    \begin{tabular}{lcc}
    \toprule
    \textbf{Setting} & \textbf{Frozen} & \textbf{Tunable} \\ \midrule
       {Image Only} & 0.52 & 0.51 \\ 
       {Image w Descriptors} & 0.96 & - \\
       {Image Only (Random IDs)} & 0.48 & 0.52 \\
    \bottomrule
    \end{tabular}
    \caption{Fine-tuned LLaVA-NeXT-7b performance (2D) when we keep the vision encoder frozen or tunable.}
    \label{tab:finetuned}
    \vspace{-0.1in}
\end{table}

One intuitive question is whether simply fine-tuning the VLMs can relieve their vision bottleneck.
To answer this question, we experiment with LLaVA-NeXT-7b and fine-tune it on the training set (13,500 pairs) of \name. We focus on the 2D case for the set of experiments described in this section. 
We consider three separate settings for training: {(1) training with only images as input, (2) training with images and textual descriptor values as input, and (3) the same setting as (1) with using random IDs as target labels.} {Details in experimental setup are included in Appendix~\ref{app:fine-tuning}.}

\subsection{Overall Performance}

\textbf{Fine-tuning does not help with vision-language mapping.} Table \ref{tab:finetuned} shows that the model fails to learn meaningful patterns in the data, even after training 50 epochs. However, when the textual descriptors are provided as input, the model can easily learn the function achieving 96\% accuracy (a 45\% improvement from the zero-shot setting). This shows that the model is unable to {extract the same descriptor values from the image input}. Unfreezing the encoder for fine-tuning also does not provide any significant improvement. We include a further discussion in Appendix~\ref{app:encoder-reps}. The results point to a limitation in the vision model's ability to encode the images, which we will investigate next.

\subsection{Understanding the Visual Representations}

In order to further understand the limitations of the vision component, we conduct an analysis to better understand how well different vision encoder models can differentiate between different paths in the dataset. To this end, we first apply a linear probe to see how easily distinguishable different images are. Specifically, given a pair of images, we first use the vision encoder to extract high-dimensional feature representations for both images. These features are then concatenated and passed through a simple binary classification layer (i.e., the probe). The probe is trained to predict a label of 1 if the images are the same and 0 otherwise. We experiment with various SOTA vision encoders, namely {CLIP \cite{radford2021learning} base and large varieties, LAION-OpenCLIP \cite{schuhmann2022laionb, OpenCLIP}, SigLip \cite{zhai2023sigmoid}, DINO \cite{caron2021emerging}, and DINO-v2 \cite{oquab2023dinov2}} and analyze how well their learned representations capture visual similarities and differences. {We use a set of 1,000 randomly sampled path pairs with a balanced label distribution to train the probe, {and look at whether the model can learn to distinguish between these paths}. We also perform an experiment where we fine-tune the vision encoder along with the probe. In this setting, our goal is to gauge if carefully fine-tuning a vision encoder can potentially improve the model's performance in low-level perception. Finally, in both settings, we present the average cosine similarity between distinct paths.}

\noindent\textbf{Vision encoders cannot distinguish between paths.} From Table \ref{tab:encoder_comparison}, it can be seen that vision encoder models are unable to provide representations that are significantly different for the probe to tell if they are the same. This is further supported by the high values for the average cosine similarity across all models.

\noindent\textbf{Fine-tuning the encoders on a discrimination task can help disentangle the visual representations.} 
By unfreezing the encoder weights and training them to identify whether two paths are identical, we enhance their adaptability to the task. The results in Table~\ref{tab:encoder_comparison} demonstrate that this approach effectively disentangles the learned representations, resulting in significantly improved performance and increased separability, as evidenced by the notably lower cosine similarity between non-identical paths. {The results thus imply the need for carefully fine-tuning task-specific vision encoders for path evaluation on \name.}

\section{Conclusion}

{
This work explored the use of VLMs as evaluators in path-planning scenarios. We evaluated a number of VLMs on our proposed benchmark, \name, and found that these models struggle with low-level perception. Specifically, we find that visual encoders used by SOTA models are unable to discern the differences between different paths in intricate scenarios. We hope that \name will inspire researchers to further explore ways to improve the visual capabilities of VLMs and contribute to finding better ways to incorporate foundation models for developing more flexible, robust, and scalable planning paradigms.  
}

\section*{Acknowledgments}

This project was supported by resources provided by the Office of Research Computing at George Mason University (URL: https://orc.gmu.edu) and funded in part by grants from the National Science Foundation (Award Number 2018631).

{
    \small
    \bibliographystyle{ieeenat_fullname}
    \bibliography{main}
}

\clearpage
\setcounter{page}{1}
\maketitlesupplementary
\setcounter{section}{0}

\renewcommand{\thesection}{\Alph{section}}

\section{Scenarios}\label{sec:appendix-scenarios}

A complete of the 15 scenarios used in \name are presented in Table \ref{tab:scenarios}. The descriptors that each scenario aims to optimize are listed in Table \ref{tab:scene-metrics}.

\begin{table*}[t!]
\footnotesize
\begin{tabularx}{\textwidth}{c|X}
\hline
\textbf{ID} & \textbf{Scenario} \\ 
\hline
1 & The agent navigating this maze is a large truck, so sharp turns (90 degrees or larger) are harder to make. It should also stay on a straight line (unless it is making a turn) as it is driving through heavy traffic. \\ 
\hline
2 & An autonomous firefighting robot is designed to navigate and operate within burning buildings to extinguish fires and rescue trapped individuals. It should explore as much of the area as possible, while maintaining a safe distance from the walls to avoid damage. \\ 
\hline
3 & As the vehicle is traversing a warzone, it must stay concealed from enemy operatives, making use of covers like walls and avoiding open spaces as much as possible. It should also reach its target (point 2) as quickly as possible. \\ 
\hline
4 & An autonomous drone delivering a package from point 1 to point 2 must take the shortest path possible due to limited fuel. It should also maintain a safe distance from surrounding buildings and make the path as straight as possible for stable flight. \\ 
\hline
5 & A robot has to deliver an aid package from point 1 to point 2 as quickly as possible. As the vehicle is moving through an earthquake-affected area, it is crucial to keep a safe distance from the walls at every moment to prevent damage from collapsing structures. \\ 
\hline
6 & A robot is moving through a museum where the walls contain fragile and expensive art pieces. Therefore, the robot should make sure to never get too close or touch any of the walls. It should also not take any abrupt turns to avoid startling the visitors. \\ 
\hline
7 & The agent navigating this construction site is a long articulated bus, making it difficult to maneuver sharp turns (90 degrees or larger). \\ 
\hline
8 & The agent navigating this trail is a wide agricultural combine harvester, making it difficult to see obstacles; hence it's hard to avoid them if they're too close. \\ 
\hline
9 & The agent navigating this busy warehouse is a long forklift, making it difficult to make sharp and abrupt turns. It should also maintain a safe distance from the obstacles at all times. \\ 
\hline
10 & The agent navigating this complex construction site is a crane with a long boom, which makes maneuvering sharp turns and around narrow passages very challenging. \\ 
\hline
11 & An autonomous taxi is navigating through an urban environment. As it is navigating heavy traffic, it should make as few sharp turns as possible and keep a safe distance from its surroundings. It should also ensure passenger comfort and safety by making left/right turns as smooth as possible. \\ 
\hline
12 & A Mars rover is exploring a Martian terrain from point 1 to point 2. The rover should conserve energy by taking the shortest path possible and avoiding unnecessary turns. Sharp turns (\( > 90 \) degrees) require higher levels of fuel and put a strain on the navigation system. \\ 
\hline
13 & An autonomous vehicle is guiding a visually impaired individual through a shopping mall. It should drive in a straight path and not make any sudden or sharp turns to ensure the individual's safety and comfort. It should also maintain a safe distance from the surrounding walls. \\ 
\hline
14 & An autonomous soil monitoring robot is tasked with navigating agricultural fields and collecting detailed soil health data. It should cover as much of the area as possible and get as close to the walls as possible to read the sensors that record the data needed. \\ 
\hline
15 & An autonomous inspection robot is tasked with navigating a nuclear power plant to inspect for radiation leaks and structural integrity. The robot has to inspect as many sections of the power plant as possible in one mission. It should get as close as possible to the walls to be able to detect minor leaks or cracks. In order to avoid accidents, it should take the straightest path possible and not make any sudden or sharp turns. \\ 
\hline
\end{tabularx}
\caption{Scenario Descriptions with Corresponding IDs}\label{tab:scenarios}
\end{table*}

\begin{table*}[t]
\centering
\small 
\resizebox{\textwidth}{!}{ 
\begin{tabular}{cccccccc}
\hline
\textbf{Scenario} & \textbf{Avg. clearance} & \textbf{Min. clearance} & \textbf{Max. clearance} & \textbf{Path length} & \textbf{\#of Sharp turns} & \textbf{Max angle} & \textbf{Smoothness} \\ 
\hline
1 & - & - & - & - & $\downarrow$ & $\downarrow$ & $\downarrow$ \\
\hline
2 & $\uparrow$ & $\uparrow$ & - & $\uparrow$ & - & - & - \\
\hline
3 & $\downarrow$ & - & $\downarrow$ & $\downarrow$ & - & - & - \\
\hline
4 & $\uparrow$ & $\uparrow$ & - & $\downarrow$ & $\downarrow$ & $\downarrow$ & $\downarrow$ \\
\hline
5 & - & $\uparrow$ & - & $\downarrow$ & - & - & - \\
\hline
6 & - & $\uparrow$ & - & - & $\downarrow$ & - & $\downarrow$ \\
\hline
7 & - & - & - & - & $\downarrow$ & $\downarrow$ & $\downarrow$ \\
\hline
8 & $\uparrow$ & $\uparrow$ & - & - & - & - & - \\
\hline
9 & - & $\uparrow$ & - & - & $\downarrow$ & $\downarrow$ & $\downarrow$ \\
\hline
10 & - & - & - & - & $\downarrow$ & $\downarrow$ & $\downarrow$ \\
\hline
11 & $\uparrow$ & $\uparrow$ & - & - & $\downarrow$ & $\downarrow$ & $\downarrow$ \\
\hline
12 & - & - & - & $\downarrow$ & $\downarrow$ & - & $\downarrow$ \\
\hline
13 & $\uparrow$ & $\uparrow$ & - & - & $\downarrow$ & $\downarrow$ & $\downarrow$ \\
\hline
14 & $\downarrow$ & - & $\downarrow$ & $\uparrow$ & - & - & - \\
\hline
15 & $\downarrow$ & - & $\downarrow$ & $\uparrow$ & $\downarrow$ & $\downarrow$ & $\downarrow$ \\
\hline
\end{tabular}
}

\caption{Scenarios and the descriptors over which they optimize. \( \downarrow \) indicates that values need to be minimized, while \( \uparrow \) means that values have to be maximized.}
\label{tab:scene-metrics}
\end{table*}

\section{Path Descriptors}\label{app:descriptors}

For a path \( P = \{p_1, p_2, .., p_n\}\) consisting of a sequence of $n$ locations $p$'s, we define each of the descriptors as follows, where {${D}(\cdot, \cdot)$} denotes the Euclidean distance between the two operands. {The three clearance metrics and Path Length share the same measuring unit, i.e., one grid size; Smoothness and Maximum angle are measured by degree; and Number of Sharp Turns is an integer count.}

\begin{enumerate}
    \item \textbf{\emph{Minimum Clearance}} measures the smallest distance between any point on the path and the nearest obstacle.
\[
\text{Min Clearance} = \min_{p_j \in P} \left( \min_{O_i \in \mathcal{O}} {D}(p_j, O_i) \right)
\]

\item \textbf{\emph{Maximum Clearance}} measures the largest distance between any point on the path and the nearest obstacle.
\[
\text{Max Clearance} = \max_{p_j \in P} \left( \min_{O_i \in \mathcal{O}} {D}(p_j, O_i) \right)
\]

\item \textbf{\emph{Average Clearance}} computes the average distance between all points on the path and the nearest obstacle.
\[
\text{Avg. Clearance} = \frac{1}{n} \sum_{p_j \in P} \left( \min_{O_i \in \mathcal{O}} {D}(p_j, O_i) \right)
\]

\item \textbf{\emph{Path Length}} sums up
the Euclidean distances between consecutive points on the path.
\[
\text{Path Length} = \sum_{j=2}^{n} {D}(p_{j-1}, p_j)
\]

\item \textbf{{\emph{Smoothness}}} is defined as the sum of the angles between consecutive segments of the path, measuring how smoothly the path changes direction.

\[
\text{Smoothness} = \sum_{j=3}^{n} \theta_j
\]
where \(\theta_j\) is the angle between the vectors \(\overrightarrow{p_{j-2}p_{j-1}}\) and \(\overrightarrow{p_{j-1}p_j}\).

\item \textbf{\emph{Number of Sharp Turns}} counts the number of turns in the path where the angle between consecutive segments exceeds 90 degrees.
\[
\text{Sharp Turns} = \sum_{j=3}^{n} \delta_j, \text{where }
\delta_j =
\begin{cases}
1 & \text{if } \theta_j > 90^\circ \\
0 & \text{otherwise}
\end{cases}
\]

\item \textbf{\emph{Maximum Angle}} denotes the largest angle between any two consecutive segments of the path.

\[
\text{Maximum angle} = \max_{j=3}^{n} \theta_j
\]

\end{enumerate}

\section{Experimental Setup}\label{app:experimental-setup}

\subsection{Prompts}\label{app:prompts}

Below, we show the different prompts used for our experiments, using Scenario \#1 in Table~\ref{tab:scenarios} as an example.
\newtcolorbox{promptbox}[1]{
    colback=cvprblue!5,    
    colframe=cvprblue!75!black,  
    coltitle=white,    
    fonttitle=\bfseries,
    title=#1,  
    boxrule=0.5pt,     
    width=\linewidth,  
    sharp corners,     
    left=5pt,         
    right=5pt,        
    top=5pt,           
    bottom=5pt,
    breakable
}
\subsubsection{Prompt for \name}
\begin{promptbox}{Prompt for \name}
    \textbf{The agent navigating this maze is a large truck, so sharp turns (90 degrees or larger) are harder to make. It should also stay on a straight line (unless it is making a turn) as it is driving through heavy traffic.} Which path better achieves the task? Path 1 is on the left side and Path 2 is on the right side. Your answer should follow the format below:\\
    \textbf{Answer}: Path 1 or Path 2.\\
    \textbf{Explanation}: Why you chose the path (1 or 2).
    
\end{promptbox}

\subsubsection{Prompt for \name w/ Descriptors}
\begin{promptbox}{Prompt for \name w/ Descriptors}
\textbf{The agent navigating this construction site is a long articulated bus, making it difficult to maneuver sharp turns (90 degrees or larger).} Which path better achieves the task? Path 1 is on the left side and Path 2 is on the right side. The following path descriptor values are computed for each path:\\

\textbf{Minimum Clearance:} The minimum distance from the obstacles. \\
\textbf{Maximum Clearance:} The maximum distance from the obstacles. \\
\textbf{Smoothness:} The sum of absolute angles between path segments. Smoother paths have a lower smoothness value. \\
\textbf{Number of sharp turns:} Number of turns that are \(>\) 90 degrees. \\
\textbf{Maximum turn angle:} The sharpest turn angle in the path. \\
\textbf{Path length:} The sum of Euclidean distances between points in the path. \\

Here are path descriptor values for Path 1:\\
\textbf{Minimum clearance:} 0.7044694115091165, \textbf{Maximum clearance:} 6.142489571740198, \textbf{Average clearance:} 3.014227976325727, \\
\textbf{Path length:} 137.5945426777758, \textbf{Smoothness:} 98.5529683186464, \textbf{Sharp turns:} 0, \textbf{Maximum angle:} 59.600200981198626. \\

Here are path descriptor values for Path 2:\\
\textbf{Minimum clearance:} 1.342789990448996, \textbf{Maximum clearance:} 5.343046965766502, \textbf{Average clearance:} 2.697627901518315, \\
\textbf{Path length:} 152.13523628046815, \textbf{Smoothness:} 480.5614409019347, \textbf{Sharp turns:} 1, \textbf{Maximum angle:} 108.15343849689171. \\

Your answer should follow the format below:\\
\textbf{Answer}: Path 1 or Path 2.\\
\textbf{Explanation}: Why you chose the path (1 or 2).
\end{promptbox}

\subsubsection{Prompt for Attribute Abstraction}\label{app:prompt-attribute-abstraction}
In the following, we present the prompt we used to understand a VLM's attribute abstraction capability (Appendix~\ref{app:level1-analysis}). The experiment queries each VLM to decide which descriptors are important for the given scenario.
\begin{promptbox}{Prompt for Attribute Abstraction}
\textbf{The agent navigating this maze is a large truck, so sharp turns (90 degrees or larger) are harder to make. It should also stay on a straight line (unless it is making a turn) as it is driving through heavy traffic.}

The following descriptors are available:
\begin{enumerate}
    \item \textbf{Minimum Clearance}: The minimum distance from the obstacles.
    \item \textbf{Maximum Clearance}: The maximum distance from the obstacles.
    \item \textbf{Average Clearance}: The average distance from the obstacles.
    \item \textbf{Smoothness}: The sum of absolute angles between path segments. Smoother paths have a lower smoothness value.
    \item \textbf{Number of Sharp Turns}: The number of turns that are $>$90 degrees.
    \item \textbf{Maximum Turn Angle}: The sharpest turn angle in the path.
    \item \textbf{Path Length}: The sum of Euclidean distances between points in the path.
\end{enumerate}

Which ones are the most important for the specified scenario?
\\
\textbf{Your answer should follow this format:}

\textbf{Answer}: list of required descriptors separated by a semicolon (;). \\
\textbf{Explanation}: Why these descriptors are important.

\end{promptbox}

\subsubsection{Prompt for Fine-grained Visual Perception}
\begin{promptbox}{Prompt for Fine-grained Visual Perception}

\textbf{Smoothness is defined as a measure of how gradual the agent's path is, minimizing sharp or abrupt changes in direction. It is calculated as the sum of angles between consecutive points (segments).}

\textbf{The task is to determine which path results in a numerically smaller value for smoothness. A smaller smoothness value means that the path has fewer abrupt turns and is smoother overall.}

Path 1 is on the left side and Path 2 is on the right side.

\textbf{Your answer should follow this format:}
\begin{itemize}
    \item \textbf{Answer}: Path 1 or Path 2.
    \item \textbf{Explanation}: Briefly explain why you chose the path (e.g., ``Path 1 has a smaller value for the given metric'').
\end{itemize}

\end{promptbox}

\subsection{Fine-grained Vision Dataset Construction} \label{app:gen-finegrained}

Intuitively, when the difference between two paths increases, it should become easier for a VLM to pick the one with a lower value. To create such a controlled dataset, for each descriptor \( m \) we consider three threshold values \( \epsilon_1, \epsilon_2, \epsilon_3 \), where \( \epsilon_1 < \epsilon_2 < \epsilon_3 \), and then sample 50 path pairs \( \{(P_i, P_j)\} \) for each threshold \( \epsilon_i \) such that \( |m(P_i) - m(P_j)| > \epsilon_i \). We list the threshold values in Table~\ref{tab:thresholds-app} and report the resulting average descriptor value for each subset in Table~\ref{tab:avg-tab}.

\begin{table}[H]
    \centering
    \footnotesize

    \begin{tabular}{lccc}
        \toprule
    \textbf{Descriptor} & \textbf{\( \epsilon_1 \)} & \textbf{\( \epsilon_2 \)} & \textbf{\( \epsilon_3 \)} \\ \midrule
    Min. Clearance & 1.0&2.0&3.0  \\ 
    Max. Clearance & 1.0&2.0&3.0   \\ 
    Avg Clearance & 1.0&2.5&5.0 \\ 
    Path Length &  50&75&100 \\ 
    Smoothness &  100&200&300 \\ 
    \# of Sharp Turns & 1&2&3 \\ 
    Max. Angle & 30&60&90 \\ 
        \bottomrule
    \end{tabular}
    \caption{Fine-grained visual perception thresholds used in our analysis for understanding a VLM's low-level perception.}
    \label{tab:thresholds-app}
\end{table}

\begin{table}[h!]

    \centering
    \footnotesize
    \begin{tabular}{lccc}
    \toprule
    \textbf{Descriptor} & \textbf{Avg. 1} & \textbf{Avg. 2} & \textbf{Avg. 3} \\ 
    \midrule
    Min. Clearance & 1.7&2.7&3.7  \\ 
    Max. Clearance & 2.4 & 3.3 & 3.8   \\ 
    Avg Clearance & 2.4 & 4.2 & 5.8 \\ 
    Path Length &  72.4 & 91.2 & 113.4 \\ 
    Smoothness &  339.4 & 372.5 & 447.6 \\ 
    \# of Sharp Turns & 2.6 & 3.5 & 4.3 \\ 
    Max. Angle & 57.6 & 74.7 & 98.7 \\ 
    \bottomrule
    \end{tabular}
    \caption{{Average differences in descriptor values of path pairs in our synthesized dataset for understanding a VLM's low-level perception.}
    }
    \label{tab:avg-tab}
\end{table}

For comparison, we also show the average difference between path pairs on \name for each descriptor in Table~\ref{tab:average-diff-benchmark}.

\begin{table}[h!]

    \centering
    \footnotesize
    \begin{tabular}{lc}
    \toprule
    \textbf{Descriptor} & \textbf{Avg Difference}  \\ 
    \midrule
    Minimum clearance & 0.593 \\
    Maximum clearance & 0.124 \\
    Average clearance & 1.31 \\ 
    Path length & 19.72 \\ 
    Smoothness & 140.93 \\ 
    \# of Sharp turns & 0.582 \\
    Maximum angle & 32.31 \\
    \bottomrule
    \end{tabular}
    \caption{{Average differences in \name.}}
    \vspace{-2mm}
    \label{tab:average-diff-benchmark}
\end{table}

\subsection{Experimental setup for VLM fine-Tuning}\label{app:fine-tuning}
In Table~\ref{tab:Hyperparams}, we present the hyper-parameters used in fine-tuning LLaVA-NeXT-7b. Our experiments were performed using A100.80gb GPUs. 
\begin{table}[H]
    \centering
    \footnotesize

    \begin{tabular}{lc}
        \toprule
     \textbf{Language Decoder} & Vicuna-7b \cite{vicuna2023} \\
     \textbf{Vision Encoder} & clip-vit-large-patch14-336 \\
     \textbf{Number of Epochs} & 50 \\
     \textbf{Batch size} & 16 \\
     \textbf{Learning Rate} & 2e-5  \\
        \bottomrule
    \end{tabular}
    \caption{Hyper-parameters used for fine-tuning.}
    \vspace{-2mm}
    \label{tab:Hyperparams}
\end{table}

\section{Additional Analysis}\label{app:additional-analysis}

\subsection{Can VLMs identify relevant descriptors?}\label{app:level1-analysis} We assess whether VLMs are able to identify the relevant descriptors for each scenario. Accordingly, we prompt the models with each scenario and the list of available descriptors and ask the model to identify which ones are important for the scenario (using the prompt template in Appendix~\ref{app:prompt-attribute-abstraction}). We run this process 5 times for each of the 15 scenarios. Table~\ref{tab:level1} shows the average number of times \emph{at least one of the} relevant descriptors are chosen by each model across all instances. (For the path pairs in \name, knowing one of the critical descriptors and being able to identify the path with a smaller or larger value as desired are sufficient for succeeding in the task.)

\begin{table}[H]
    \centering
    \footnotesize
    \begin{tabular}{lc}
        \toprule
        \textbf{Model} & \textbf{Avg. Success}  \\
        \midrule
        \textbf{GPT-4o-mini} & 1.00  \\
        \textbf{GPT-4o} & 1.00 \\
        \midrule
        \textbf{LLaVa-NeXT-7b} & 1.00 \\
        \textbf{Qwen2-VL-7b} & 1.00 \\
        \textbf{LLaVA-OneVision-7b} & 1.00 \\
        \textbf{Intern-VL2-8b} & 0.92 \\
        \textbf{LLaMa-3.2-11b} &  0.95 \\
        \textbf{LLaVa-NeXT-13b} & 0.93 \\
        \bottomrule
    \end{tabular}
    \caption{Average number of times \emph{at least one of the} relevant descriptors are chosen by each model across all scenarios.}
    \vspace{-2mm}
    \label{tab:level1}
\end{table}

\subsection{Some decision-making scenarios are easier than others.}\label{app:gpt4o-breakdown} 
\begin{figure}[tb]
    \centering
    \includegraphics[width=\linewidth]{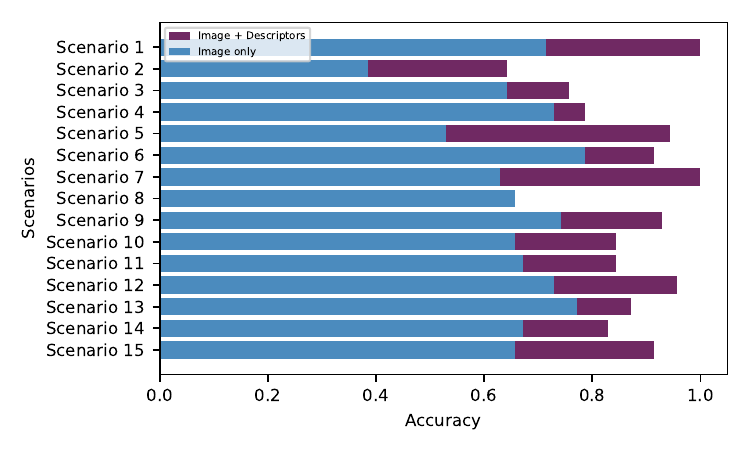} 
    \caption{GPT-4o performance per scenario (2D)}\label{fig:perf-per-scenario}
\end{figure}

In other to get a better understanding of how the models utilize the provided information, we look at the performance gain in each of the 15 scenarios for the best-performing model (GPT-4o). This performance is broken down in Figure~\ref{fig:perf-per-scenario}. We notice that GPT-4o may \emph{overuse its commonsense}. Specifically, we observed that the model particularly struggled with Scenario \#2 (Table~\ref{tab:scenarios}). This is because, contrary to most real-case scenarios, this case requires maximizing the path length, to be able to cover as much of the area as possible. However, as finding the shortest path is generally a more common command, GPT-4o has likely been pre-trained to prioritize this pattern and thus often opts for paths minimizing the length on \name.
On the other hand, GPT-4o was also shown to \emph{struggle in using its commonsense knowledge properly}.
Specifically, its performance on Scenario \#8 does not improve even when the descriptors are listed. This is because, in this scenario, the specification does not explicitly list the aspects that the model has to consider. This indicates that the models struggle to use their commonsense reasoning to identify aspects that are not explicitly stated in the specifications.

\subsection{Representations produced by the vision encoders are ambiguous.}\label{app:encoder-reps} 
One challenge encountered with encoders such as CLIP occurs is that similar images are represented almost identically \cite{tong2024eyeswideshutexploring, shengbang2023massproducing}. To investigate whether this is the issue faced in our case, we examine whether the model can easily memorize a small set of samples when fine-tuned for 100 epochs. If the representations were fully unambiguous, this memorization task should be straightforward. The results, displayed in Table~\ref{tab:memorization}, indicate that the model struggles to differentiate between several image and scenario pairs, as reflected by its low performance in the standard setting. Notably, performance significantly improves when descriptors or even random tokens are added to the language descriptions, making each image-language pair unique. 

\begin{table}[H]
    \footnotesize
    \centering
    \resizebox{\linewidth}{!}{
    \begin{tabular}{lccc}
    \toprule
    \textbf{\# of Instances} & \textbf{Image Only} & \textbf{Image + Desc.} & \textbf{Image + Rand. Tokens} \\ \midrule
       50 & 0.72 & 1.0 & 1.0  \\ 
       100 & 0.68 & 1.0 & 1.0 \\
       200 & 0.77 & 1.0 & 1.0 \\
       \bottomrule
    \end{tabular}
    }
    \caption{Fine-tuned LLaVA-NeXT-7b (w/ frozen vision encoder)'s accuracy in memorizing small training samples (2D) when each path is represented by only its image, image with descriptor values, or image with random text tokens (used to validate the VLM's small-data memorization capability).}
    \label{tab:memorization}
\end{table}

\section{Failure Examples}\label{app:error-analysis}

\begin{figure*}[tb]
\centering
\begin{tcolorbox}[colback=cvprblue!5!white, colframe=cvprblue!75!black,  title=Failure Examples on \name]
    \centering
    \begin{minipage}{\linewidth}
        \centering
        \frame{\includegraphics[width=0.5\textwidth]{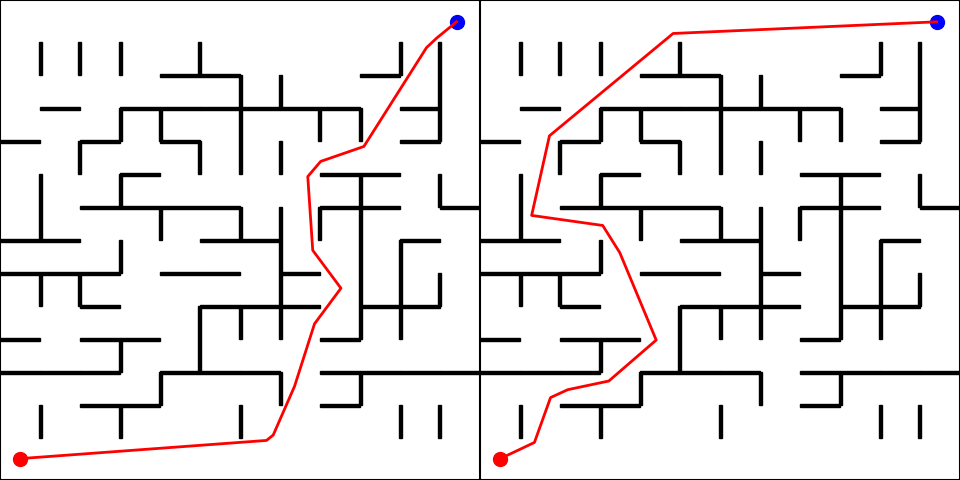}} 
    \end{minipage}
    \begin{minipage}{\textwidth}
        \footnotesize
        \rule{\textwidth}{0.2pt}  

        \textbf{Prompt} \\ \textit{
The agent navigating this maze is a large truck, so sharp turns (90 degrees or larger) are harder to make. It should also stay on a straight line (unless it is making a turn) as it is driving through heavy traffic. Which path better achieves the task? Path 1 is on the left side and Path 2 is on the right side.\\
Your answer should follow the format below:\\
\textbf{Answer: Path 1 or Path 2.}\\
\textbf{Explanation: Why you chose the path (1 or 2).}}\\
        \rule{\textwidth}{0.4pt}  \\
        \textbf{GPT-4o}\\
        \textbf{Answer}: \textbf{\textcolor{red}{Path 2}} \\
        \textbf{Explanation}: \textbf{\textcolor{red}{Path 2 has fewer sharp turns compared to Path 1}}, making it more suitable for a large truck that has difficulty making sharp turns. Additionally, Path 2 allows the truck to stay on a straight line for longer distances, which is beneficial for navigating through heavy traffic. \\

        \textbf{LLaMA-11b}\\
        \textbf{Answer}: \textbf{\textcolor{red}{Path 2}} \\
        \textbf{Explanation}: 
        
        Step 1: Identify the key constraints of the task.\\The agent is a large truck, which means it has difficulty making sharp turns (90 degrees or larger). Additionally, the agent should stay on a straight line as much as possible due to heavy traffic.\\Step 2: Analyze Path 1.\\Path 1 involves several sharp turns, including a 90-degree turn. This would be challenging for the large truck to navigate.\\nStep 3: Analyze Path 2.\\Path 2 involves \textbf{\textcolor{red}{fewer sharp turns and more gradual turns}}, which would be easier for the large truck to navigate.\\Step 4: Determine the best path based on the constraints.\\Given the constraints, Path 2 is the better option as it involves fewer sharp turns and more gradual turns, making it easier for the large truck to navigate.\\

        \textbf{LLaVA-OneVision-7b} \\
    \textbf{Answer}: \textbf{\textcolor{red}{Path 2}} \\
        \textbf{Explanation}:
        \textbf{\textcolor{red}{Path 2 is a straight line from the starting point to the end point}}, which is the most efficient route for a large truck with difficulty making sharp turns. It avoids any sharp turns and stays on a straight line, which is ideal for driving through heavy traffic.
    \end{minipage}
\end{tcolorbox}
\caption{Examples of model failure on \name.}
\label{fig:failures-nodesc}
\end{figure*}

\begin{figure*}[b]
\centering
\begin{tcolorbox}[colback=cvprblue!5!white, colframe=cvprblue!75!black, title=Failure Examples on \name w/ Descriptors]

    \centering
    \begin{minipage}{0.6\textwidth}
        \centering
        \includegraphics[width=\linewidth]{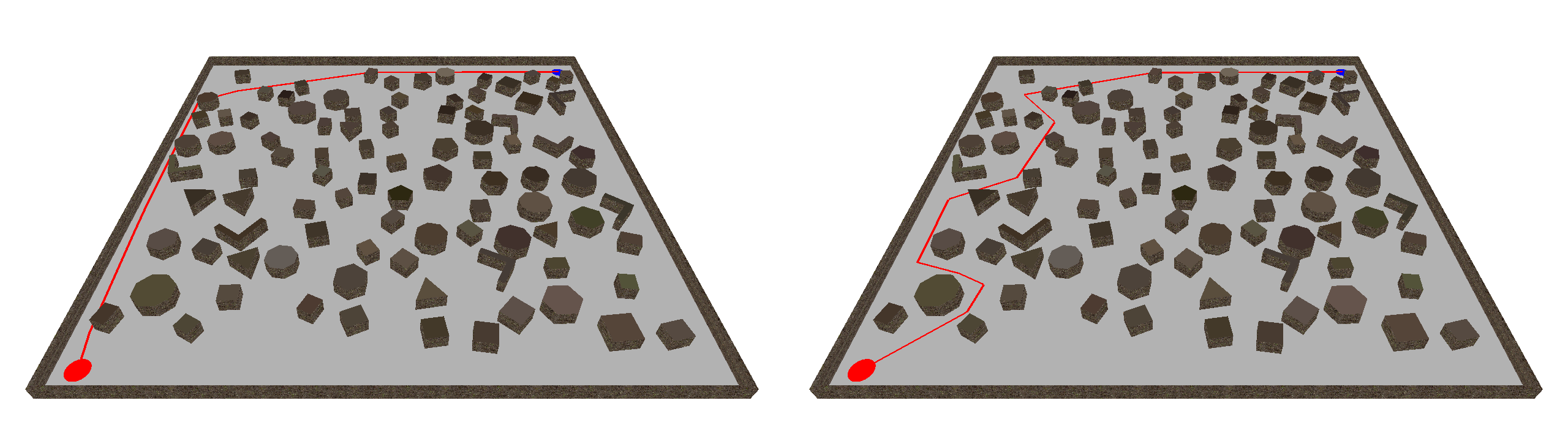} 
    \end{minipage}
    \begin{minipage}{\textwidth}
        \footnotesize
        \rule{\textwidth}{0.4pt}  

        \textbf{Prompt} \\ \textit{
The agent navigating this construction site is a long articulated bus, making it difficult to maneuver sharp turns (90 degrees or larger). Which path better achieves the task? Path 1 is on the left side and Path 2 is on the right side. The following path descriptor values are computed for each path:\\
\textbf{Minimum Clearance:} The minimum distance from the obstacles. \\
\textbf{Maximum Clearance:} The maximum distance from the obstacles. \\
\textbf{Smoothness:} The sum of absolute angles between path segments. Smoother paths have a lower smoothness value.\\
\textbf{Number of sharp turns:} Number of turns that are \(>\) 90 degrees.\\
\textbf{Maximum turn angle:} The sharpest turn angle in the path.\\
\textbf{Path length:} The sum of Euclidean distances between points in the path.\\
Here are path descriptor values for Path 1: \\
\textbf{Minimum clearance:} 0.7044694115091165, \textbf{Maximum clearance:} 6.142489571740198, \textbf{Average clearance:} 3.014227976325727, \textbf{Path length:} 137.5945426777758, \textbf{Smoothness:} 98.5529683186464, \textbf{Sharp turns:} 0, \textbf{Maximum angle:} 59.600200981198626.\\
Here are path descriptor values for Path 2: \\
\textbf{Minimum clearance:} 1.342789990448996, \textbf{Maximum clearance:} 5.343046965766502, \textbf{Average clearance:} 2.697627901518315, \textbf{Path length:} 152.13523628046815, \textbf{Smoothness:} 480.5614409019347, \textbf{Sharp turns:} 1, \textbf{Maximum angle:} 108.15343849689171.\\
Your answer should follow the format below:\\
\textbf{Answer: Path 1 or Path 2.}\\
\textbf{Explanation: Why you chose the path (1 or 2).}}\\
        \rule{\textwidth}{0.4pt}  \\
        \textbf{LLaVA-NeXT-7b}\\
        \textbf{Answer}: \textbf{\textcolor{red}{Path 2}} \\
        \textbf{Explanation}: Path 2 is chosen because it has a \textbf{\textcolor{red}{lower smoothness value}}, which indicates that it has \textbf{\textcolor{red}{fewer sharp turns}} and a more gradual path. This is advantageous for a long articulated bus, as it would be easier to maneuver and avoid obstacles without getting stuck or damaging the bus. \textbf{\textcolor{red}{The higher clearance values for Path 2}} also suggest that it is less likely to encounter issues with obstacles. \\
        
        \textbf{LLaVA-NeXT-13b} \\
        \textbf{Answer}: \textbf{\textcolor{red}{Path 2}} \\
        \textbf{Explanation:} Path 2 is chosen because it has a \textbf{\textcolor{red}{lower smoothness value}}, which indicates that it is a smoother path. This is important for a long articulated bus, as it will require less maneuvering and turning to navigate the construction site. The path length is also slightly shorter, which could be advantageous in terms of fuel efficiency and ease of navigation. Additionally, Path 2 has a higher minimum clearance, which is important for the bus to avoid hitting any obstacles. The maximum clearance is also higher, providing more margin for error. The only downside is that it has one more sharp turn compared to Path 1, but given the other advantages, it is still a more suitable path for the articulated bus. \\
        
        \textbf{InternVL2-8b} \\
        \textbf{Answer}: \textbf{\textcolor{red}{Path 2}} \\
        \textbf{Explanation}: Path 1 has a higher number of sharp turns (0 compared to 1) and a higher smoothness value (98.5529683186464 compared to 480.5614409019347), which suggests that it is a smoother path overall. However, Path 1 also has lower clearance values compared to Path 2. Moreover, Path 1 has a smaller path length (137.5945426777758 compared to 152.13523628046815), which indicates a shorter path that still manages to navigate through the obstacles effectively. On the contrary, Path 2, although it doesn't have as many sharp turns, has better parameters like Maximum Clearance (5.343046965766502), Average clearance (2.697627901518315) which are indicators for the agent like a long articulated bus to maneuver safely. \textbf{\textcolor{red}{(Wrong descriptors are chosen)}}

    \end{minipage}
\end{tcolorbox}
\caption{Examples of model failure on \name when prompted with w/ descriptor values.}
\label{fig:example-llava-desc}
\end{figure*}

We present some examples of failures produced by different models in Figures \ref{fig:failures-nodesc} and \ref{fig:example-llava-desc}.

\section{Human Performance on the Task}

We sampled a set of 60 examples (2 from each scenario in both 2D and 3D) from PathEval and tasked four annotators with selecting the better path. Two of the annotators are authors of the paper (Ann. 3\&4) and the other two are externals unfamiliar with the work (Ann. 1\&2).  
We notice that in all cases the human annotators significantly outperform GPT-4o on the sampled subset, however, we notice that in the 3D case, the task can be difficult even for humans.

\begin{table}[htbp]
\vspace{-0.12in}
    \centering
    \resizebox{\columnwidth}{!}{%
    \begin{tabular}{ccccc}
        \hline 
        \textbf{Ann. 1} & \textbf{Ann. 2} & \textbf{Ann. 3} & \textbf{Ann. 4} & \textbf{GPT-4o} \\
        \hline
        0.90/0.70 & 0.73/0.60 & 0.90/0.76 & 0.80/0.73 & 0.53/0.47 \\
        \hline
    \end{tabular}%
    }
\vspace{-0.12in}
\end{table}


\end{document}